%% file: main.tex
\documentclass[10pt,twocolumn,letterpaper]{article}

\usepackage{cvpr}              

\usepackage{graphicx}
\usepackage{amsmath}
\usepackage{amssymb}
\usepackage{booktabs}
\usepackage{tikz}
\usepackage{pgfplots}
\pgfplotsset{compat=newest}
\usepackage{subfiles}

\usepackage[pagebackref,breaklinks,colorlinks]{hyperref}

\usepackage[capitalize]{cleveref}
\crefname{section}{Sec.}{Secs.}
\Crefname{section}{Section}{Sections}
\Crefname{table}{Table}{Tables}
\crefname{table}{Tab.}{Tabs.}

\def\cvprPaperID{*****} 
\def\confName{CVPR}
\def\confYear{2023}

\begin{document}

\title{ThinResNet: A New Baseline for Structured Convolutional Networks Pruning}

\author{{\small Hugo Tessier}\\
{\small IMT Atlantique}\\
{\small Lab-STICC, UMR CNRS 6285}\\{\small 29238 Brest, France}\\
{\tt\footnotesize hugo.tessier@imt-atlantique.fr}
\and
{\small Ghouti Boukli Hacene}\\
{\small Sony Europe}\\
{\small R\&D Center, Stuttgart Laboratory 1}\\
{\small 70327 Stuttgart, Germany}\\
{\tt\footnotesize ghouti.bouklihacene@sony.com}
\and
{\small Vincent Gripon}\\
{\small IMT Atlantique}\\
{\small Lab-STICC, UMR CNRS 6285}\\{\small 29238 Brest, France}\\
{\tt\footnotesize vincent.gripon@imt-atlantique.fr}
}
\maketitle

\begin{abstract}
   Pruning is a compression method which aims to improve the efficiency of neural networks by reducing their number of parameters while maintaining a good performance, thus enhancing the performance-to-cost ratio in nontrivial ways. Of particular interest are structured pruning techniques, in which whole portions of parameters are removed altogether, resulting in easier to leverage shrunk architectures.
   Since its growth in popularity in the recent years, pruning gave birth to countless papers and contributions, resulting first in critical inconsistencies in the way results are compared, and then to a collective effort to establish standardized benchmarks.
   However, said benchmarks are based on training practices that date from several years ago and do not align with current practices.
   In this work, we verify how results in the recent literature of pruning hold up against networks that underwent both state-of-the-art training methods and trivial model scaling.
   We find that the latter clearly and utterly outperform all the literature we compared to, proving that updating standard pruning benchmarks and re-evaluating classical methods in their light is an absolute necessity.
   We thus introduce a new challenging baseline to compare structured pruning to: ThinResNet.
\end{abstract}

\begin{figure*}
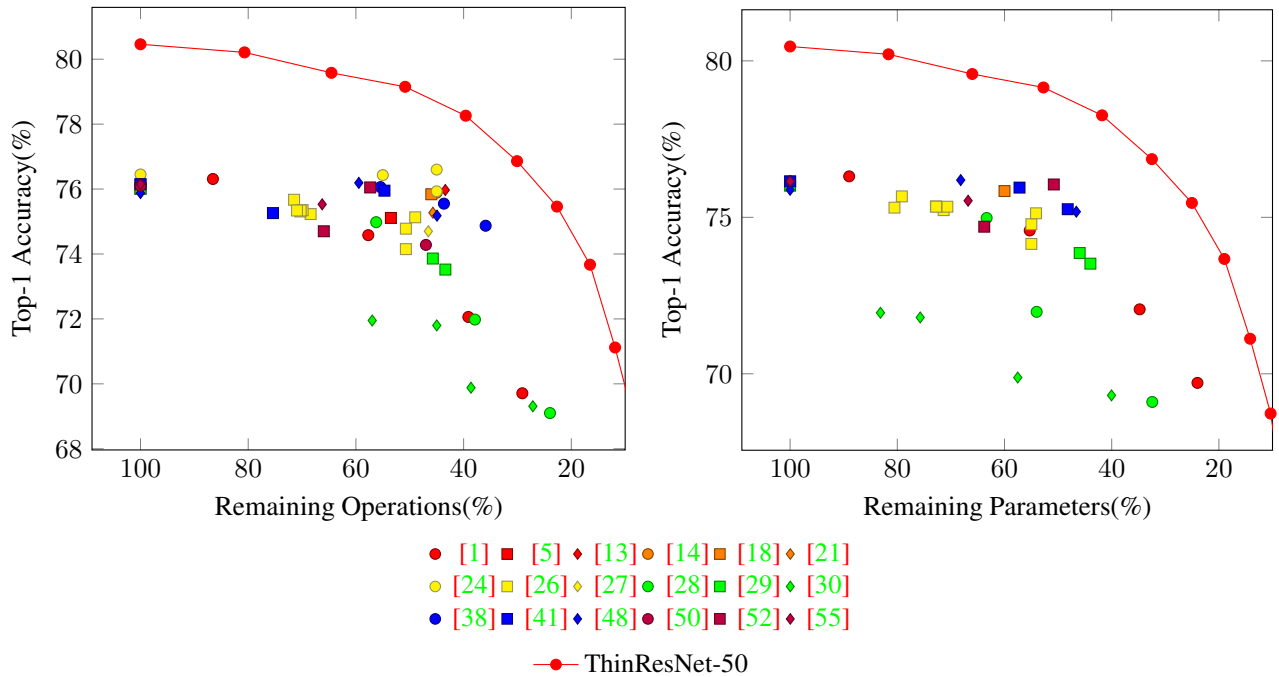

    \centering
    \resizebox{0.49\linewidth}{!}{
    \subfile{figures/operations_imagenet}
    }
    \resizebox{0.49\linewidth}{!}{
    \subfile{figures/parameters_imagenet}
    }
        \ref{operations_imagenet_legend}

        \ref{parameters_imagenet_legend}

    \caption{Results for ResNet-50 on the ImageNet ILSVRC2012 dataset.}
    \label{fig:imagenet_results}
\end{figure*}

\begin{figure*}
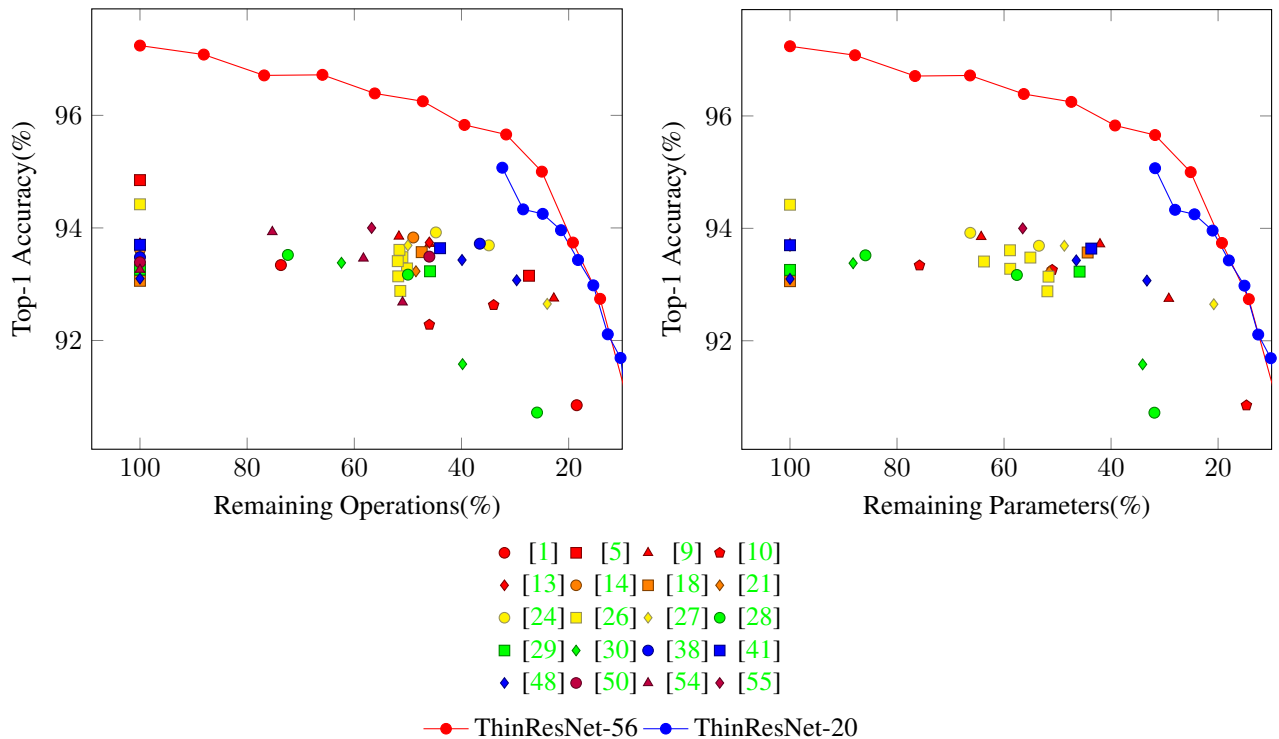

    \centering
    \resizebox{0.49\linewidth}{!}{
    \subfile{figures/operations_cifar}
    }
    \resizebox{0.49\linewidth}{!}{
    \subfile{figures/parameters_cifar}
    }
        \ref{operations_cifar_legend}
        
        \ref{parameters_cifar_legend}
    \caption{Results on CIFAR-10. All reference methods use ResNet-56. We provide results for both ThinResNet-56 and ThinResNet-20 to compare the influence of width and depth. The ratio of remaining operations and parameters of ThinResNet-20 is shown while keeping ResNet-56 as the reference, so that a regular ResNet-20 of width 16 is shown as having virtually undergone pruning.}
    \label{fig:cifar_results}
\end{figure*}

\section{Introduction}
\label{sec:intro}

Being the state of the art in countless domains, such as computer vision~\cite{chen2023symbolic}, language processing~\cite{brown2020language} or image generation~\cite{rombach2022high}, deep neural networks remain computationally expensive for both learning and inference, which tends to make them unsuitable for many applications, as well as environmentally unfriendly.
This is the reason why a large literature emerged to compress such networks; their goal is usually to maximize a performance-to-cost ratio, where the cost can account for computations, latency, bandwidth, memory, energy usage...
Among them, the field of neural networks pruning~\cite{han2015learning} is particularly active, and counts numerous publications each year.
Structured pruning~\cite{li2016pruning}, where whole portions of the networks are removed altogether, is especially promising, as it leads to more reliable cost reduction compared to non-structured methods.

However, some papers have raised concerns about how pruning methods compare to each others: lack of consistency between benchmarks\cite{blalock2020state}, techniques that may not work well on actual hardware~\cite{ma2021non}, or even unexpected side-effects that make it counter-productive~\cite{tessier2022energy}, pruning has had many problems in the past, that many papers in the literature have endeavored to solve.

Globally, the field of pruning suffers from not being built on proper theoretical foundations, as well as from missing reliable benchmarks. More precisely, theoretical foundations are based on wild assumptions, such as using tailor expansions, which only hold for small perturbations whereas pruning methods typically nullify a lot of parameters in considered architectures, and only hold for making sure the performance right after pruning remains high, while most if not all of pruning methods include training steps after removing parameters.
Consequently, it is not absurd to question the effectiveness of proposed methods, which is often taken for granted in the literature.
This is why we decided, as a first step, to propose a new, fresh baseline which to compare structured pruning to, based on popular architecture/dataset pairs in the literature (\textit{i.e.} ResNet-50~\cite{he2016deep}/ImageNet ILSVRC2012~\cite{russakovsky2015imagenet} and ResNet-56/CIFAR-10~\cite{krizhevsky2009learning}).
This new baseline is both simple and challenging, in order for future pruning methods to showcase their true efficiency.

We therefore propose ThinResNet, that are simply ResNets with a uniformly reduced proportion of channels in each layer (\textit{a.k.a.} ``width''), but trained using state-of-the-art methods.
Such reduced architectures are not only trivial to generate, but also contain no hidden cost~\cite{tessier2022leveraging}.
We also release trained models on CIFAR10, CIFAR100 and ImageNet ILSVRC2012, for various target flops and number of parameters, that can be directly used for fine tuning, feature extraction or direct classification on considered classes.

We compare our baseline to a wide array of methods from the literature, that are usually tested on outdated training procedures, in order to showcase the order of magnitude of the gap between these benchmarks and what is actually possible with modern training methods.
We not only state that such a comparison is not unfair, the ultimate goal of pruning staying to maximize the performance-to-cost ratio, but also necessary, because what is measured for a certain baseline performance may not translate well once using more modern training procedures.

We therefore encourage future works to use our ThinResNet as their main point of comparison, as long as no better performing methods are available.
We made both our code and pretrained models available to download on our Github page: \url{https://github.com/brain-bzh/baseline-pruning}.

To summarize, our contributions are the following:
\begin{itemize}
    \item We introduce a simple baseline for structured pruning methods, for image classification with ResNets,
    \item We show that this baseline clearly outperform results from existing literature, and we describe how benchmarks in the field should be updated, for both practical and theoretical reasons,
    \item We provide the code and the trained models on our Github, including a gradient of ThinResNet ranging from 4Gflops (25M parameters) down to 48Mflops (570k parameters) pretrained on ImageNet ILSVRC2012 and ready to be fine-tuned or deployed for edge applications.
\end{itemize}

\section{Background}

Compression of neural networks generated a lot of interest during the last decade, and many types of compression methods have been introduced, such as quantization~\cite{courbariaux2015binaryconnect}, distillation~\cite{hinton2015distilling}, clustering~\cite{soulie2016compression} or the one we focus on here: pruning, or the removal of parts of neural networks.

Initially born in the late 80's~\cite{lecun1989optimal}, pruning grew in popularity with the work of Han~\textit{et~al.}~\cite{han2015learning} in 2015.
Since then, many different methods have been published, which can mainly be divided into two different categories: non-structured and structured pruning.

Non-structured pruning simply involves pruning isolated parameters without consideration for the particular layout of the pruning masks and how easy it will be to leverage them on hardware.
This type of pruning has two main advantages: 1) its simplicity makes room for more experimental and theoretical experimentation~\cite{han2015learning,frankle2018lottery,molchanov2017variational} and 2) its fine-grained nature allows more easily to reach extremely high pruning rates~\cite{tessier2022rethinking}, which is especially interesting since some libraries are able to leverage non-structured sparsity only at the condition of exceeding, for example, 95\% of zeros in the case of cuSPARSE~\cite{yao2019balanced}.
Another, more marginal advantage, is that the introduced redundancy in values among parameters allows for a better compression through, for example, Huffman coding~\cite{han2015deep}, which can be interesting for some low-power devices~\cite{younes2022inter}.

Structured pruning, instead, prunes larger type of structures, such as whole channels, which has numerous advantages: 1) since it results in a reduction in the architecture itself, any framework and hardware can leverage it, 2) it does not only reduce the number of parameters but also the operations required, as well as the size of intermediate representations, thus reducing the memory utilization during inference, 3) there are enough channels in a network to allow for a reasonably fine-grained and efficient pruning, compared to layers or blocks.
These significant advantages made this type of pruning (``channel'' or ``filter'' pruning, loosely referred as ``structured'' pruning as it became the standard by default) especially popular very quickly in the literature~\cite{li2016pruning,liu2017learning}, and the topic of interest for our paper.

Whatever the type of pruning, another aspect has to be taken account of: its distribution across layers.
Indeed, while pruning parameters or filters, it is possible either to set a same criterion on all layers at once, which is called ``global'' pruning~\cite{han2015learning,liu2017learning}, or to set manually a given ``local'' pruning rate on each layer independently, whether it is the same rate (``uniform'') or not (``non-uniform'')~\cite{li2016pruning}.
Theoretically, global pruning should bring better results, as it may not restrict itself to the predefined architectures fixed by local, manual pruning, which could be suboptimal.

However, some contributions have raised doubts about the efficiency of pruning against more trivial baselines trained from scratch~\cite{liu2018rethinking,gale2019state}, or about its ability to produce efficient architectures~\cite{tessier2022leveraging,tessier2022energy}.
Finally, in 2020, Blalock~\textit{et~al.}~\cite{blalock2020state} raised the alarm about the consistency of benchmarks in the literature, warning about how few papers actually bothered to compare on the same pairs of architectures and datasets and under the same training conditions: everything remained to be standardized.
They therefore proposed ShrinkBench\footnote{https://github.com/JJGO/shrinkbench}, to help such an effort, which revealed fruitful as many papers use it nowadays.
Since then, two pairs became very widespread for comparison between methods: ResNet-56/CIFAR-10, performing between 93.5\% and 94.5\% Top-1 accuracy and ResNet-50/ImageNet ILSVRC2012, performing around 76.15\% Top-1 accuracy (cf. Table~\ref{tab:cifar_table} and \ref{tab:imagenet_table}).
It is debatable whether these choices are the best to test the effectiveness of pruning methods, which is beyond the scope of this paper.

\section{Methodology}\label{sec:methodology}

Our goal is to confront the literature of structured pruning with what modern training methods are able to provide as the simplest baseline possible, \textit{i.e.} networks reduced from the start and trained from scratch.
Not only is such a method reminiscent of the concerns of Liu~\textit{et~al.}~\cite{liu2018rethinking} and Gale~\textit{et~al.}~\cite{gale2019state}, but it also has the merit of being unequivocal, as we propose the simplest way possible to emulate the behavior of local and uniform channel pruning.

\paragraph{Training conditions}

We used the same hyperparameters as those shown on the Pytorch blog\footnote{\url{https://pytorch.org/blog/how-to-train-state-of-the-art-models-using-torchvision-latest-primitives/}}.
Even though they were designed for training on ImageNet ILSVRC2012, we found that they delivered comparable improvements when training on CIFAR-10.
Here is a brief summary of the training conditions we used:
\begin{itemize}
    \item Batch Size: 1024
    \item Steps: 750,000, adding 5 epochs of warmup
    \item Optimizer: SGD, with a weight decay of $2\times10^{-5}$ (except for batch-normalization layers that are spared) and a momentum of $0.9$
    \item Scheduler: linear during the warmup, then cosine annealing
    \item Learning rate: $5\times10^{-3}$ at the beginning, then $0.5$ at the end of the warmup, then decreases until it reaches 0
    \item Criterion: cross-entropy loss, with a label smoothing~\cite{szegedy2016rethinking} factor of $0.1$
    \item Data Augmentation:
    \begin{itemize}
        \item TrivialAugment~\cite{muller2021trivialaugment}
        \item Random Erasing~\cite{zhong2020random, devries2017improved}, with a probability of 0.1
        \item Mixup~\cite{zhang2017mixup} with a Beta(0.2, 0.2) distribution
        \item Cutmix~\cite{yun2019cutmix} with a Beta(1.0, 1.0) distribution
        \item FixRes~\cite{touvron2019fixing} mitigation, reducing the resolution during training from $224\times224$ to $176\times176$
        \item Exponential Moving Average (EMA), every 32 steps, with a decay of $0.99998$
    \end{itemize}
    \item Inference Resize: during validation, input images are resized from $224\times224$ to $232\times232$
    \item Interpolation: bilinear
\end{itemize}
However, we do not apply Repeated Augmentation~\cite{hoffer2019augment, berman2019multigrain}, as we found that it brings too little improvement for a prohibitive cost in term of training time.

\paragraph{Replicability}\label{sec:replic}

We share our code and provide the trained models whose results are reported in Table~\ref{tab:ours_structured_table}
. We ran each experiment under deterministic conditions, using the same seed 0. We did not report the best accuracy across epochs, but only that of the EMA model at the very last epoch, even when it is not the best result available.

\paragraph{Measurements}

While counting the number of remaining parameters can seem trivial, the case of batch-normalization layers can be ambiguous, as they are eventually fused with convolution layers after training. However, since counting their weights or not has a negligible impact on the ratio of remaining parameters, we kept them.
Concerning the number of operations, we used the torchinfo\footnote{\url{https://github.com/TylerYep/torchinfo}} utility, that gave us counts of ``mult-adds'' operations (a.k.a. MACs), even though most papers report counts of FLOPs. Since FLOPs are usually almost exactly twice the number of MACs, and since we report pruning ratios instead of raw numbers, we consider the difference between the two to be negligible.

\paragraph{Pseudo-Pruning Strategies}

We simply varied the width (\textit{i.e.} the number of channels) of each convolutional layer in the architecture proportionally, which emulates the resulting architecture of a strictly uniform local structured pruning strategy.
Besides its simplicity, this method has a great advantage: it brings no discrepancies at all between layers, as can be the case when performing pruning~\cite{tessier2022leveraging}.

\paragraph{Choice of the ResNet Variant}

Although the architecture of ResNet-50 is absolutely standardized, there are actually two competing variants for ResNets dedicated to CIFAR-10, such as ResNet-56 and ResNet-20: their shortcut modules can contain either a $1\times1$ convolution layer or a padding operation.
The padding variant is the one used in ShrinkBench (sourced from another popular repository\footnote{\url{https://github.com/akamaster/pytorch_resnet_cifar10}}).
However, these padding operations are much more constraining when doing any kind of structured pruning that is not uniform, because of the channel dependencies between layers~\cite{tessier2022leveraging}.
Even though this did not pose problem with our own pseudo-pruning strategy, it may actually be a significant burden to the rest of the literature.
After having verified that replacing the paddings by the other variant did not impact the performance at all, while increasing the number of parameters by less than 1\%, we decided to use the variant using convolutions in order to advocate for its use over that of the padding one.

\section{Results}

\paragraph{Choice of reference methods}



The methods we chose to compare to were selected among the publications of various international conferences.
Were kept those which matched the following criteria: 1) the papers provide results for at least ResNet-56/CIFAR-10 or ResNet-50/ImageNet, 2) they provide their results in the form of a table, since it is almost impossible to obtain exact values otherwise (e.g. from a graph or a figure), 3) the absolute accuracy of each result is provided (or, at least, possible to infer from a baseline), and 4) the pruning rate, in term of operations or parameters, is indicated or possible to infer.
Each of these criteria, that are essential to extract exploitable results from the papers without having to reproduce the experiments, ended up excluding a large portion of papers, so that our final selection only represent a minority of the papers we reviewed.

\paragraph{Structured Pruning on ImageNet ILSVRC2012}

Figure~\ref{fig:imagenet_results} compare results from the literature of structured pruning to our own, for ResNet-50 trained on ImageNet ILSVRC2012.
Results from the reference methods are directly extracted from the original papers (as for all figures in this article).
Since multiple papers report results in term of operations but not parameters, some references are missing for the curve showing the ratio of remaining parameters.
Detailed results of reference methods are reported in Table~\ref{tab:imagenet_table}, and our own results are shown in Table~\ref{tab:ours_structured_table}.

\paragraph{Structured Pruning on CIFAR-10}

Figure~\ref{fig:cifar_results} compare results from the literature of structured pruning to our own, for both ResNet-56 and ResNet-20, trained on CIFAR-10.
The reason why we tested two different architectures was to distinguish the impact of width and depth on the efficiency of the architecture and whether the two reduction strategies would lead to different results.
Some references are also missing when showing remaining parameters.
Detailed results of reference methods are reported in Table~\ref{tab:cifar_table}, and our own results are shown in Table~\ref{tab:ours_structured_table}.



\section{Discussions}

\subsection{A New Baseline}

As can be seen in Figure~\ref{fig:imagenet_results} and~\ref{fig:cifar_results}, our results for ResNet-50, ResNet-56 and ResNet-20 beat their counterparts from the literature by a wide margin.
Since these points were obtained using the simplest method possible, while providing networks that are actually reduced without any ambiguity about the efficiency of their implementation on hardware, they can serve as a relevant baseline to beat for future pruning methods.
This is why we provide both the code, to obtain them, and the networks themselves to download on the dedicated Github page.

\subsection{The Importance of Proper Training}

The implications of our work are not only practical, but also theoretical.
Indeed, the question is not just ``what are the best available compressed networks ?'' but also ``are the current benchmarks reliable, to tell apart methods that work or not ?''.
As mentioned by Tessier~\textit{et~al.}~\cite{tessier2023convolutional}, an insufficient post-pruning retraining can lead to erroneous conclusions.
Therefore, even though our results are obviously insufficient to tell that pruning does not work, it evidences that, in the absence of a proper, up-to-date benchmark, it is not possible to tell that the previous results would scale the same way once properly retrained.

\subsection{New Benchmark Paradigms}

All of this means that sticking to out-of-date training conditions, for the sake of making comparison easier, may not turn out to be such a good idea.
Indeed, not only is the accuracy-to-operations ratio (or, more generally, performance-to-cost ratio) not the only thing that matters in the end, from an applicative point of view, but, as we mentioned, old benchmarks can create erroneous conclusions.
Therefore, even though some criteria need to be fixed, such as the performance and cost metrics, the way to present results and make code available, and the datasets on which to compare methods, everything else should likely not be fixed, and instead be kept the most up-to-date possible.

This also means that, ultimately, we should likely not stick to simple ResNets, as we did ourself (even though we considered testing other architectures to be out of the scope of this work).
Indeed, why should we prune them, if other architectures are available while being more efficient than even compressed ResNets, especially if we consider pruning as a subcategory of Neural Architecture Search~\cite{tessier2023convolutional}?
However, keeping at least one reference network is still useful, not only because it still helps making methods easier to compare, but also because it can serve as a sanity check: if a method looks to work well for a given custom architecture but not at all on ResNet-50/ImageNet, then there may be some overlooked variables explaining the good results.

\section{Conclusion}

In this contribution, we propose to compare results from the literature of structured pruning to a very simple baseline method that uses modern training hyperparameters.
It turns out that our results largely outmatch all the references we compared to, and shows that the literature needs to update its benchmarks.
We therefore made our compressed networks available to download, as well as our code, to serve as a new baseline to beat for future pruning methods.
We encourage the literature not only to compare to our results for future methods, but also to renew previous experiments to verify if conclusions that were to be drawn from old benchmarks do scale well to new ones, or if all previous discussions in the literature were unfortunately just a byproduct of insufficient retraining methods.

{\small
\bibliographystyle{ieee_fullname}
\bibliography{egbib}
}

\clearpage

\appendix

\begin{table*}[h]
    \centering
    \subfile{tables/resnet50_ours}

\subfile{tables/resnet56_ours}

\subfile{tables/resnet20_ours}
    \caption{Our results for the pairs ThinResNet-50/ImageNet ILSVRC2012, ThinResNet-56/CIFAR-10 and ThinResNet-20/CIFAR-10. We report the width (the number of channels in the initial embedding), the Top-1 accuracy (Acc.) in percents, the percentage of remaining operations (R.O.) as well as that of remaining parameters (R.P.). 
    We also report the latency (Lat.) in milliseconds.
These values were measured using a script that we provide in our Github repository.
The measurements were performed on Intel Xeon Silver 4208 CPU, as doing so gave more steady results compared to GPU.
Each experiment involved running inferences during 1000 seconds, and we report the average of all the  points generated during this duration.
    }
    \label{tab:ours_structured_table}
\end{table*}


\begin{table*}[h]
    \centering
    \subfile{tables/cifar_table}

\subfile{tables/cifar_table_bis}

\subfile{tables/cifar_table_ter}
    \caption{Results from the literature of structured pruning, for ResNet-56 on CIFAR-10. Values are directly extracted from the original papers. We report the width (the number of channels in the initial embedding), the Top-1 accuracy (Acc.) in percents, the percentage of remaining operations (R.O.) as well as that of remaining parameters (R.P.).}
    \label{tab:cifar_table}
\end{table*}

\begin{table*}[h]
    \centering
    \subfile{tables/imagenet_table}

\subfile{tables/imagenet_table_bis}

\subfile{tables/imagenet_table_ter}
    \caption{Results from the literature of structured pruning, for ResNet-50 on ImageNet ILSVRC2012. Values are directly extracted from the original papers. We report the width (the number of channels in the initial embedding), the Top-1 accuracy (Acc.) in percents, the percentage of remaining operations (R.O.) as well as that of remaining parameters (R.P.).}
    \label{tab:imagenet_table}
\end{table*}


\end{document}

%% file: figures/operations_imagenet.tex
\begin{tikzpicture}
    \begin{axis}[
    xlabel=Remaining Operations(\%),
    ylabel=Top-1 Accuracy(\%),
    x dir=reverse,
    xmin=10,
    legend columns=6, 
    legend to name=operations_imagenet_legend,
    legend style={draw=none},
    legend entries={
\cite{alwani2022decore}, 
\cite{cai2022prior}, 
\cite{ganjdanesh2022interpretations}, 
\cite{gao2022disentangled}, 
\cite{he2022filter}, 
\cite{kang2020operation}, 
\cite{lee2022ensemble}, 
\cite{li2022revisiting}, 
\cite{li2020group}, 
\cite{lin2020hrank}, 
\cite{lin2020channel}, 
\cite{lin2019towards}, 
\cite{shang2022neural}, 
\cite{tang2020scop}, 
\cite{you2019gate}, 
\cite{yu2022topology}, 
\cite{yvinec2022singe}, 
\cite{zhong2021revisit}, 
}
    ]

\addplot [only marks, mark=*, mark options={fill=red, draw=red!50!black}] coordinates {
(100,76.15)
(86.55256723716381,76.31)
(57.701711491442545,74.58)
(39.119804400978,72.06)
(29.095354523227385,69.71)
}; 

\addplot [only marks, mark=square*, mark options={fill=red, draw=red!50!black}] coordinates {
(100,76.01)
(53.5,75.11)
}; 

\addplot [only marks, mark=diamond*, mark options={fill=red, draw=red!50!black}] coordinates {
(100,76.13)
(43.4,75.97)
}; 

\addplot [only marks, mark=*, mark options={fill=orange, draw=orange!50!black}] coordinates {
(100,76.13)
(45.0,75.89)
}; 

\addplot [only marks, mark=square*, mark options={fill=orange, draw=orange!50!black}] coordinates {
(100,76.13)
(46,75.84)
}; 

\addplot [only marks, mark=diamond*, mark options={fill=orange, draw=orange!50!black}] coordinates {
(100,75.89)
(45.7,75.27)
}; 

\addplot [only marks, mark=*, mark options={fill=yellow, draw=yellow!50!black}] coordinates {
(100,76.45)
(55.0,76.43)
(45.0,75.93)
(45.0,76.6)
}; 

\addplot [only marks, mark=square*, mark options={fill=yellow, draw=yellow!50!black}] coordinates {
(100,76.15)
(68.41,75.23)
(71.48,75.67)
(69.89,75.35)
(70.32,75.31)
(70.26,75.34)
(70.94,75.34)
(50.72,74.15)
(50.72,74.78)
(48.99,75.13)
}; 

\addplot [only marks, mark=diamond*, mark options={fill=yellow, draw=yellow!50!black}] coordinates {
(46.55,74.7)
}; 

\addplot [only marks, mark=*, mark options={fill=green, draw=green!50!black}] coordinates {
(100,76.15)
(56.23471882640586,74.98)
(37.89731051344744,71.98)
(23.96088019559902,69.1)
}; 

\addplot [only marks, mark=square*, mark options={fill=green, draw=green!50!black}] coordinates {
(100,76.01)
(43.38926904756148,73.52)
(45.7141475445511,73.86)
}; 

\addplot [only marks, mark=diamond*, mark options={fill=green, draw=green!50!black}] coordinates {
(100,76.15)
(56.96821515892421,71.95)
(38.63080684596577,69.88)
(44.987775061124694,71.8)
(27.139364303178485,69.31)
}; 

\addplot [only marks, mark=*, mark options={fill=blue, draw=blue!50!black}] coordinates {
(100,76.13)
(55.44,76.06)
(43.65,75.55)
(35.91,74.87)
}; 

\addplot [only marks, mark=square*, mark options={fill=blue, draw=blue!50!black}] coordinates {
(100,76.15)
(54.7,75.95)
(75.4,75.26)
}; 

\addplot [only marks, mark=diamond*, mark options={fill=blue, draw=blue!50!black}] coordinates {
(100,75.88)
(59.46,76.19)
(44.94,75.18)
}; 

\addplot [only marks, mark=*, mark options={fill=purple, draw=purple!50!black}] coordinates {
(100,76.1)
(47,74.28)
}; 

\addplot [only marks, mark=square*, mark options={fill=purple, draw=purple!50!black}] coordinates {
(57.35,76.05)
(65.96,74.7)
}; 

\addplot [only marks, mark=diamond*, mark options={fill=purple, draw=purple!50!black}] coordinates {
(100,76.15)
(66.25999999999999,75.53)
}; 


\addplot [mark=*,red,width=2pt] coordinates { 
(100.0,80.46)
(80.68459657701712,80.21)
(64.54767726161369,79.58)
(50.8557457212714,79.15)
(39.60880195599022,78.26)
(30.073349633251834,76.86)
(22.665036674816626,75.46)
(16.5281173594132,73.67)
(11.8737817499, 71.12)
(8.166259168704157,68.73)
(5.50122249388753,65.39)
(3.4963325183374083,59.53)
(2.13310078747,53.34)
(1.1809290953545233,42.72)
(0.59967362489, 29.12)
};

    \end{axis}
\end{tikzpicture}


%% file: figures/parameters_imagenet.tex
\begin{tikzpicture}
    \begin{axis}[
    xlabel=Remaining Parameters(\%),
    ylabel=Top-1 Accuracy(\%),
    x dir=reverse,
    xmin=10,
    legend columns=5, 
    legend to name=parameters_imagenet_legend,
    legend style={draw=none, text=black},
    legend entries={
ThinResNet-50
}
    ]

\addplot [only marks, mark=*, mark options={fill=red, draw=red!50!black},forget plot] coordinates {
(100,76.15)
(88.98039215686275,76.31)
(55.294117647058826,74.58)
(34.78431372549019,72.06)
(24.0,69.71)
};

\addplot [only marks, mark=square*, mark options={fill=orange, draw=orange!50!black},forget plot] coordinates {
(100,76.13)
(60,75.84)
};
    
\addplot [only marks, mark=square*, mark options={fill=yellow, draw=yellow!50!black},forget plot] coordinates {
(100,76.15)
(71.35,75.23)
(79.15,75.67)
(72.8,75.35)
(80.53,75.31)
(70.66,75.34)
(72.76,75.34)
(54.99,74.15)
(54.99,74.78)
(54.12,75.13)
};

\addplot [only marks, mark=*, mark options={fill=green, draw=green!50!black},forget plot] coordinates {
(100,76.15)
(63.33333333333333,74.98)
(54.0,71.98)
(32.431372549019606,69.1)
};

\addplot [only marks, mark=square*, mark options={fill=green, draw=green!50!black},forget plot] coordinates {
(100,76.01)
(43.97496087636933,73.52)
(45.97026604068858,73.86)
};

\addplot [only marks, mark=diamond*, mark options={fill=green, draw=green!50!black},forget plot] coordinates {
(100,76.15)
(83.13725490196077,71.95)
(57.529411764705884,69.88)
(75.72549019607843,71.8)
(40.03921568627452,69.31)
};

\addplot [only marks, mark=square*, mark options={fill=blue, draw=blue!50!black},forget plot] coordinates {
(100,76.15)
(57.2,75.95)
(48.2,75.26)
};

\addplot [only marks, mark=diamond*, mark options={fill=blue, draw=blue!50!black},forget plot] coordinates {
(100,75.88)
(68.17,76.19)
(46.6,75.18)
};

\addplot [only marks, mark=square*, mark options={fill=purple, draw=purple!50!black},forget plot] coordinates {
(50.8,76.05)
(63.78,74.7)
};

\addplot [only marks, mark=diamond*, mark options={fill=purple, draw=purple!50!black},forget plot] coordinates {
(100,76.15)
(66.78999999999999,75.53)
};


\addplot [mark=*,red,width=2pt] coordinates { 
(100.0,80.46)
(81.640625,80.21)
(66.015625,79.58)
(52.734375,79.15)
(41.796875,78.26)
(32.5,76.86)
(25.039062499999996,75.46)
(18.984375,73.67)
(14.1720681807,71.12)
(10.3515625,68.73)
(7.421875,65.39)
(5.15625,59.53)
(3.47968418242,53.34)
(2.21484375,42.72)
(1.28133423318,29.12)
};

    \end{axis}
\end{tikzpicture}


%% file: figures/operations_cifar.tex
\begin{tikzpicture}
    \begin{axis}[
    xlabel=Remaining Operations(\%),
    ylabel=Top-1 Accuracy(\%),
    x dir=reverse,
    xmin=10,
    legend columns=4, 
    legend to name=operations_cifar_legend,
    legend style={draw=none, text=black},
    legend entries={
\cite{alwani2022decore},
\cite{cai2022prior},
\cite{di2022channel}, 
\cite{elkerdawy2022fire},
\cite{ganjdanesh2022interpretations},
\cite{gao2022disentangled},
\cite{he2022filter},
\cite{kang2020operation},
\cite{lee2022ensemble},
\cite{li2022revisiting},
\cite{li2020group},
\cite{lin2020hrank},
\cite{lin2020channel},
\cite{lin2019towards},
\cite{shang2022neural},
\cite{tang2020scop},
\cite{you2019gate},
\cite{yu2022topology},
\cite{zhang2022weighted},
\cite{zhong2021revisit},
}
    ]

\addplot [only marks, mark=*, mark options={fill=red, draw=red!50!black}] coordinates {
(100,93.26)
(73.7,93.34)
(50.1,93.26)
(18.5,90.85)
};

\addplot [only marks, mark=square*, mark options={fill=red, draw=red!50!black}] coordinates {
(100,94.85)
(27.400000000000006,93.15)
};

\addplot [only marks, mark=triangle*, mark options={fill=red, draw=red!50!black}] coordinates {
(100,93.72)
(51.68,93.85)
(45.95,93.72)
(22.77,92.75)
}; 

\addplot [only marks, mark=pentagon*, mark options={fill=red, draw=red!50!black}] coordinates {
(34,92.63)
(46,92.28)
}; 

\addplot [only marks, mark=diamond*, mark options={fill=red, draw=red!50!black}] coordinates {
(100,93.56)
(46.0,93.74)
};

\addplot [only marks, mark=*, mark options={fill=orange, draw=orange!50!black}] coordinates {
(100,93.62)
(49.0,93.83)
};

\addplot [only marks, mark=square*, mark options={fill=orange, draw=orange!50!black}] coordinates {
(100,93.06)
(47.42,93.57)
};

\addplot [only marks, mark=diamond*, mark options={fill=orange, draw=orange!50!black}] coordinates {
(100,93.69)
(48.5,93.23)
};

\addplot [only marks, mark=*, mark options={fill=yellow, draw=yellow!50!black}] coordinates {
(44.78,93.92)
(34.89,93.69)
};

\addplot [only marks, mark=square*, mark options={fill=yellow, draw=yellow!50!black}] coordinates {
(100,94.42)
(50.16,93.28)
(51.03,93.48)
(51.58,93.61)
(51.81,93.14)
(51.89,93.41)
(51.42,92.88)
};

\addplot [only marks, mark=diamond*, mark options={fill=yellow, draw=yellow!50!black}] coordinates {
(50,93.69)
(24,92.65)
};

\addplot [only marks, mark=*, mark options={fill=green, draw=green!50!black}] coordinates {
(100,93.26)
(72.4,93.52)
(50,93.17)
(25.900000000000006,90.72)
};

\addplot [only marks, mark=square*, mark options={fill=green, draw=green!50!black}] coordinates {
(100,93.26)
(45.87055320482683,93.23)
};

\addplot [only marks, mark=diamond*, mark options={fill=green, draw=green!50!black}] coordinates {
(62.4,93.38)
(39.8,91.58)
};

\addplot [only marks, mark=*, mark options={fill=blue, draw=blue!50!black}] coordinates {
(100,93.48)
(36.58,93.72)
};

\addplot [only marks, mark=square*, mark options={fill=blue, draw=blue!50!black}] coordinates {
(100,93.7)
(44,93.64)
};

\addplot [only marks, mark=diamond*, mark options={fill=blue, draw=blue!50!black}] coordinates {
(100,93.1)
(39.9,93.42999999999999)
(29.700000000000003,93.07)
};

\addplot [only marks, mark=*, mark options={fill=purple, draw=purple!50!black}] coordinates {
(100,93.39)
(46,93.49)
};

\addplot [only marks, mark=triangle*, mark options={fill=purple, draw=purple!50!black}] coordinates {
(100,93.26)
(75.3,93.93)
(58.3,93.46)
(51.0,92.68)
}; 

\addplot [only marks, mark=diamond*, mark options={fill=purple, draw=purple!50!black}] coordinates {
(56.77,94.0)
};


\addplot [mark=*,red,width=2pt] coordinates { 
(100.0,97.24)
(88.09523809523809,97.08)
(76.82539682539684,96.71)
(65.95238095238095,96.72)
(56.19047619047619,96.39)
(47.22222222222222,96.25)
(39.44444444444444,95.83)
(31.666666666666664,95.66)
(25.0,95.0)
(19.20634920634921,93.74)
(14.126984126984127,92.74)
(9.84,91.23)
(6.301587301587301,88.88)
(3.5634920634920637,84.57)
(1.5952380952380953,76.36)
(0.41031746031746036,57.19)
};


\addplot [mark=*,blue,width=2pt] coordinates { 
(32.38095238095238,95.07)
(28.49206349206349,94.33)
(24.841269841269842,94.25)
(21.428571428571427,93.96)
(18.253968253968253,93.43)
(15.396825396825397,92.98)
(12.698412698412698,92.11)
(10.317460317460316,91.69)
(8.174603174603174,90.3)
(6.2857142857142865,88.92)
(4.634920634920634,87.16)
(3.2380952380952377,84.18)
(2.0873015873015874,81.43)
(1.1904761904761905,76.77)
(0.5444444444444444,66.12)
(0.14682539682539683,47.33)
};

    \end{axis}
\end{tikzpicture}


%% file: figures/parameters_cifar.tex
\begin{tikzpicture}
    \begin{axis}[
    xlabel=Remaining Parameters(\%),
    ylabel=Top-1 Accuracy(\%),
    x dir=reverse,
    xmin = 10,
    legend columns=2, 
    legend to name=parameters_cifar_legend,
    legend style={draw=none, text=black},
    legend entries={ThinResNet-56, ThinResNet-20}
    ]

\addplot [only marks, mark=triangle*, mark options={fill=red, draw=red!50!black},forget plot] coordinates {
(100,93.72)
(64.27,93.85)
(42.07,93.72)
(29.21,92.75)
}; 

\addplot [only marks, mark=pentagon*, mark options={fill=red, draw=red!50!black},forget plot] coordinates {
(100,93.26)
(75.8,93.34)
(51,93.26)
(14.700000000000003,90.85)
}; 

\addplot [only marks, mark=square*, mark options={fill=orange, draw=orange!50!black},forget plot] coordinates {
(100,93.06)
(44.37,93.57)
}; 

\addplot [only marks, mark=diamond*, mark options={fill=orange, draw=orange!50!black},forget plot] coordinates {
(100,93.69)
(51.53,93.23)
};

\addplot [only marks, mark=*, mark options={fill=yellow, draw=yellow!50!black},forget plot] coordinates {
(66.31,93.92)
(53.48,93.69)
}; 

\addplot [only marks, mark=square*, mark options={fill=yellow, draw=yellow!50!black},forget plot] coordinates {
(100,94.42)
(58.85,93.28)
(55.08,93.48)
(58.89,93.61)
(51.69,93.14)
(63.75,93.41)
(51.9,92.88)
}; 

\addplot [only marks, mark=diamond*, mark options={fill=yellow, draw=yellow!50!black},forget plot] coordinates {
(48.73,93.69)
(20.8,92.65)
}; 

\addplot [only marks, mark=*, mark options={fill=green, draw=green!50!black},forget plot] coordinates {
(100,93.26)
(85.9,93.52)
(57.6,93.17)
(31.900000000000006,90.72)
}; 

\addplot [only marks, mark=square*, mark options={fill=green, draw=green!50!black},forget plot] coordinates {
(100,93.26)
(45.88235294117647,93.23)
}; 

\addplot [only marks, mark=diamond*, mark options={fill=green, draw=green!50!black},forget plot] coordinates {
(88.2,93.38)
(34.099999999999994,91.58)
}; 

\addplot [only marks, mark=square*, mark options={fill=blue, draw=blue!50!black},forget plot] coordinates {
(100,93.7)
(43.7,93.64)
};

\addplot [only marks, mark=diamond*, mark options={fill=blue, draw=blue!50!black},forget plot] coordinates {
(100,93.1)
(46.5,93.42999999999999)
(33.3,93.07)
}; 

\addplot [only marks, mark=diamond*, mark options={fill=purple, draw=purple!50!black},forget plot] coordinates {
(56.51,94.0)
}; 


\addplot [mark=*,red,width=2pt] coordinates { 
(100.0,97.24)
(87.85046728971963,97.08)
(76.63551401869158,96.71)
(66.35514018691589,96.72)
(56.30841121495327,96.39)
(47.429906542056074,96.25)
(39.25233644859813,95.83)
(31.775700934579437,95.66)
(25.116822429906545,95.0)
(19.27570093457944,93.74)
(14.252336448598129,92.74)
(9.89,91.23)
(6.366822429906542,88.88)
(3.6098130841121496,84.57)
(1.6355140186915886,76.36)
(0.4287383177570094,57.19)
};


\addplot [mark=*,blue,width=2pt] coordinates { 
(31.775700934579437,95.07)
(28.037383177570092,94.33)
(24.415887850467293,94.25)
(21.02803738317757,93.96)
(17.990654205607477,93.43)
(15.070093457943926,92.98)
(12.5,92.11)
(10.151869158878505,91.69)
(8.037383177570094,90.3)
(6.16822429906542,88.92)
(4.544392523364485,87.16)
(3.177570093457944,84.18)
(2.0443925233644857,81.43)
(1.167056074766355,76.77)
(0.5327102803738317,66.12)
(0.14369158878504673,47.33)
};

    \end{axis}
\end{tikzpicture}


%% file: tables/resnet50_ours.tex
\begin{tabular}[t]{ccccc}
\toprule

\multicolumn{4}{c}{ThinResNet-50/ImageNet}\\

\midrule
Width & Acc. & R.O. & R.P. & Lat.\\

64&80.46& 100.0& 100.0&181.32\\

60&80.21& 80.68& 81.64&156.23\\

56&79.58& 64.55& 66.02&135.36\\

52&79.15& 50.86& 52.73&107.80\\

48&78.26& 39.61& 41.80&87.31\\

44&76.86& 30.07& 32.50&68.38\\

40&75.46& 22.67& 25.04&57.75\\

36&73.67& 16.53& 18.98&47.41\\

32&71.12& 11.88& 14.14&34.54\\

28&68.73& 8.17& 10.35&27.57\\

24&65.39& 5.50& 7.42&22.26\\

20&59.53& 3.05& 5.16&17.17\\

16& 53.34 & 2.13 & 3.47&12.59\\

12&42.72& 1.18& 2.21&9.73\\

8& 29.12 & 0.60 & 1.208& 7.31\\


\bottomrule
\end{tabular}

%% file: tables/resnet56_ours.tex
\begin{tabular}[t]{ccccc}
\toprule
\multicolumn{4}{c}{ThinResNet-56/CIFAR-10}\\

\midrule

Width & Acc. & R.O. & R.P. & Lat. \\

16&97.24& 100.0& 100.0&12.46\\

15&97.08& 88.1& 87.85&11.57\\

14&96.71& 76.83& 76.64&10.73\\

13&96.72& 65.95& 66.36&10.21\\

12&96.39& 56.19& 56.31&9.31\\

11&96.25& 47.22& 47.43&8.51\\

10&95.83& 39.44& 39.25&7.96\\

9&95.66& 31.67& 31.78&7.56\\

8&95.0& 25.0& 25.12&6.71\\

7&93.74& 19.21& 19.28&5.99\\

6&92.74& 14.13& 14.25&5.59\\

5& 91.23 &9.84& 9.89&5.18\\

4&88.88& 6.3& 6.37&4.76\\

3&84.57& 3.56& 3.61&4.31\\

2&76.36& 1.6& 1.64&3.79\\

1&57.19& 0.41& 0.43&3.43\\

\bottomrule
\end{tabular}

%% file: tables/resnet20_ours.tex
\begin{tabular}[t]{ccccc}
\toprule
\multicolumn{4}{c}{ThinResNet-20/CIFAR-10}\\

\midrule
Width & Acc. & R.O. & R.P. & Lat. \\

16&95.07& 32.38& 31.78&4.38\\

15&94.33& 28.49& 28.04&4.12\\

14&94.25& 24.84& 24.42&3.84\\

13&93.96& 21.43& 21.03&3.56\\

12&93.43& 18.25& 17.99&3.31\\

11&92.98& 15.40 & 15.07&3.07\\

10&92.11& 12.70 & 12.50 & 2.87\\

9&91.69& 10.32& 10.15&2.66\\

8&90.3& 8.17& 8.04&2.43\\

7&88.92& 6.29& 6.17&2.23\\

6&87.16& 4.63& 4.54&2.06\\

5&84.18& 3.24& 3.18&1.89\\

4&81.43& 2.09& 2.04&1.72\\

3&76.77& 1.19& 1.17&1.57\\

2&66.12& 0.54& 0.53&1.43\\

1&47.33& 0.15& 0.14&1.28\\

\bottomrule
\end{tabular}

%% file: tables/cifar_table.tex
\begin{tabular}[t]{cccc}
\toprule
Method & Acc. & R.O. & R.P. \\

\midrule
\cite{alwani2022decore}&93.26& 100& 100\\ 

&93.34& 73.7& 75.8\\

&93.26& 50.1& 51\\

&90.85& 18.5& 14.7\\

\midrule
\cite{cai2022prior}&94.85& 100& 100\\ 

&93.15& 27.4& -\\

\midrule
\cite{di2022channel}&93.72& 100& 100\\ 

&93.85& 51.68& 64.27\\

&93.72& 45.95& 42.07\\

&92.75& 22.77& 29.21\\

\midrule
\cite{elkerdawy2022fire}&92.63& 34& -\\ 

&92.28& 46& -\\

\midrule
\cite{ganjdanesh2022interpretations}&93.56& 100& 100\\ 

&93.74& 46.0& -\\

\midrule
\cite{gao2022disentangled}&93.62& 100& 100\\ 

&93.83& 49.0& -\\

\midrule
\cite{he2022filter}&93.06& 100& 100\\ 

&93.57& 47.42& 44.37\\

\bottomrule
\end{tabular}

%% file: tables/cifar_table_bis.tex
\begin{tabular}[t]{cccc}
\toprule
Method & Acc. & R.O. & R.P. \\

\midrule
\cite{kang2020operation}&93.69& 100& 100\\ 

&93.23& 48.5& 51.53\\

\midrule
\cite{lee2022ensemble}&93.92& 44.78& 66.31\\ 

&93.69& 34.89& 53.48\\

\midrule
\cite{li2022revisiting}&94.42& 100& 100\\ 

&93.28& 50.16& 58.85\\

&93.48& 51.03& 55.08\\

&93.61& 51.58& 58.89\\

&93.14& 51.81& 51.69\\

&93.41& 51.89& 63.75\\

&92.88& 51.42& 51.9\\

\midrule
\cite{li2020group}&93.69& 50& 48.73\\ 

&92.65& 24& 20.8\\

\midrule
\cite{lin2020channel}&93.26& 100& 100\\ 

&93.23& 45.87& 45.88\\

\midrule
\cite{lin2020hrank}&93.26& 100& 100\\ 

&93.52& 72.4& 85.9\\

&93.17& 50& 57.6\\

&90.72& 25.9& 31.9\\

\bottomrule
\end{tabular}

%% file: tables/cifar_table_ter.tex
\begin{tabular}[t]{cccc}
\toprule
Method & Acc. & R.O. & R.P. \\

\midrule
\cite{lin2019towards}&93.38& 62.4& 88.2\\ 

&91.58& 39.8& 34.1\\

\midrule
\cite{shang2022neural}&93.48& 100& 100\\ 

&93.72& 36.58& -\\

\midrule
\cite{tang2020scop}&93.7& 100& 100\\ 

&93.64& 44& 43.7\\

\midrule
\cite{you2019gate}&93.1& 100& 100\\ 

&93.43& 39.9& 46.5\\

&93.07& 29.7& 33.3\\

\midrule
\cite{yu2022topology}&93.39& 100& 100\\ 

&93.49& 46& -\\

\midrule
\cite{zhang2022weighted}&93.26& 100& 100\\ 

&93.93& 75.3& -\\

&93.46& 58.3& -\\

&92.68& 51.0& -\\

\midrule
\cite{zhong2021revisit}&94.0& 56.77& 56.51\\ 

\bottomrule
\end{tabular}

%% file: tables/imagenet_table.tex
\begin{tabular}[t]{cccc}
\toprule
Method & Acc. & R.O. & R.P. \\

\midrule
\cite{alwani2022decore}&76.15&100&100\\

&76.31& 86.55& 88.98\\ 

&74.58& 57.7& 55.29\\

&72.06& 39.12& 34.78\\

&69.71& 29.1& 24.0\\

\midrule
\cite{cai2022prior}&76.01&100&100\\
&75.11& 53.5& -\\ 

\midrule
\cite{ganjdanesh2022interpretations}&76.13&100&100\\
&75.97& 43.4& -\\ 

\midrule
\cite{gao2022disentangled}&76.13&100&100\\
&75.89& 45.0& -\\ 

\midrule
\cite{he2022filter}&76.13&100&100\\
&75.84& 46& 60\\ 

\midrule
\cite{kang2020operation}&75.89&100&100\\
&75.27& 45.7& -\\ 

\midrule
\cite{lee2022ensemble}&76.45&100&100\\
&76.43& 55.0& -\\ 

&75.93& 45.0& -\\

&76.6& 45.0& -\\

\bottomrule
\end{tabular}

%% file: tables/imagenet_table_bis.tex
\begin{tabular}[t]{cccc}
\toprule
Method & Acc. & R.O. & R.P. \\

\midrule
\cite{li2022revisiting}&76.15& 100& 100\\ 

&75.23& 68.41& 71.35\\

&75.67& 71.48& 79.15\\

&75.35& 69.89& 72.8\\

&75.31& 70.32& 80.53\\

&75.34& 70.26& 70.66\\

&75.34& 70.94& 72.76\\

&74.15& 50.72& 54.99\\

&74.78& 50.72& 54.99\\

&75.13& 48.99& 54.12\\

\midrule
\cite{li2020group}&74.7& 46.55& -\\ 

\midrule
\cite{lin2020hrank}&76.15&100&100\\
&74.98& 56.23& 63.33\\ 

&71.98& 37.9& 54.0\\

&69.1& 23.96& 32.43\\

\midrule
\cite{lin2020channel}&76.01&100&100\\
&73.52& 43.39& 43.97\\ 

&73.86& 45.71& 45.97\\

\bottomrule
\end{tabular}

%% file: tables/imagenet_table_ter.tex
\begin{tabular}[t]{cccc}
\toprule
Method & Acc. & R.O. & R.P. \\

\midrule
\cite{lin2019towards}&76.15&100&100\\
&71.95& 56.97& 83.14\\ 

&69.88& 38.63& 57.53\\

&71.8& 44.99& 75.73\\

&69.31& 27.14& 40.04\\

\midrule
\cite{shang2022neural}&76.13&100&100\\
&76.06& 55.44& -\\ 

&75.55& 43.65& -\\

&74.87& 35.91& -\\

\midrule
\cite{tang2020scop}&76.15&100&100\\
&75.95& 54.7& 57.2\\ 

&75.26& 75.4& 48.2\\

\midrule
\cite{you2019gate}&75.88&100&100\\
&76.19& 59.46& 68.17\\ 

&75.18& 44.94& 46.6\\

\midrule
\cite{yu2022topology}&76.1&100&100\\
&74.28& 47& -\\ 

\midrule
\cite{yvinec2022singe}&76.05& 57.35& 50.8\\ 

&74.7& 65.96& 63.78\\

\midrule
\cite{zhong2021revisit}&76.15&100&100\\
&75.53& 66.26& 66.79\\ 

\bottomrule
\end{tabular}

%% file: main.bbl
\begin{thebibliography}{10}\itemsep=-1pt

\bibitem{alwani2022decore}
Manoj Alwani, Yang Wang, and Vashisht Madhavan.
\newblock Decore: Deep compression with reinforcement learning.
\newblock In {\em Proceedings of the IEEE/CVF Conference on Computer Vision and
  Pattern Recognition}, pages 12349--12359, 2022.

\bibitem{berman2019multigrain}
Maxim Berman, Herv{\'e} J{\'e}gou, Andrea Vedaldi, Iasonas Kokkinos, and
  Matthijs Douze.
\newblock Multigrain: a unified image embedding for classes and instances.
\newblock {\em arXiv preprint arXiv:1902.05509}, 2019.

\bibitem{blalock2020state}
Davis Blalock, Jose~Javier Gonzalez~Ortiz, Jonathan Frankle, and John Guttag.
\newblock What is the state of neural network pruning?
\newblock {\em Proceedings of machine learning and systems}, 2:129--146, 2020.

\bibitem{brown2020language}
Tom Brown, Benjamin Mann, Nick Ryder, Melanie Subbiah, Jared~D Kaplan, Prafulla
  Dhariwal, Arvind Neelakantan, Pranav Shyam, Girish Sastry, Amanda Askell,
  et~al.
\newblock Language models are few-shot learners.
\newblock {\em Advances in neural information processing systems},
  33:1877--1901, 2020.

\bibitem{cai2022prior}
Linhang Cai, Zhulin An, Chuanguang Yang, Yangchun Yan, and Yongjun Xu.
\newblock Prior gradient mask guided pruning-aware fine-tuning.
\newblock In {\em Proceedings of the AAAI Conference on Artificial
  Intelligence}, volume~36, pages 140--148, 2022.

\bibitem{chen2023symbolic}
Xiangning Chen, Chen Liang, Da Huang, Esteban Real, Kaiyuan Wang, Yao Liu, Hieu
  Pham, Xuanyi Dong, Thang Luong, Cho-Jui Hsieh, et~al.
\newblock Symbolic discovery of optimization algorithms.
\newblock {\em arXiv preprint arXiv:2302.06675}, 2023.

\bibitem{courbariaux2015binaryconnect}
Matthieu Courbariaux, Yoshua Bengio, and Jean-Pierre David.
\newblock Binaryconnect: Training deep neural networks with binary weights
  during propagations.
\newblock {\em Advances in neural information processing systems}, 28, 2015.

\bibitem{devries2017improved}
Terrance DeVries and Graham~W Taylor.
\newblock Improved regularization of convolutional neural networks with cutout.
\newblock {\em arXiv preprint arXiv:1708.04552}, 2017.

\bibitem{di2022channel}
Yuan~Cao Di~Jiang and Qiang Yang.
\newblock On the channel pruning using graph convolution network for
  convolutional neural network acceleration.
\newblock In {\em Proc. Int. Joint Conf. Artif. Intell}, volume~7, pages
  3107--3113, 2022.

\bibitem{elkerdawy2022fire}
Sara Elkerdawy, Mostafa Elhoushi, Hong Zhang, and Nilanjan Ray.
\newblock Fire together wire together: A dynamic pruning approach with
  self-supervised mask prediction.
\newblock In {\em Proceedings of the IEEE/CVF Conference on Computer Vision and
  Pattern Recognition}, pages 12454--12463, 2022.

\bibitem{frankle2018lottery}
Jonathan Frankle and Michael Carbin.
\newblock The lottery ticket hypothesis: Finding sparse, trainable neural
  networks.
\newblock {\em arXiv preprint arXiv:1803.03635}, 2018.

\bibitem{gale2019state}
Trevor Gale, Erich Elsen, and Sara Hooker.
\newblock The state of sparsity in deep neural networks.
\newblock {\em arXiv preprint arXiv:1902.09574}, 2019.

\bibitem{ganjdanesh2022interpretations}
Alireza Ganjdanesh, Shangqian Gao, and Heng Huang.
\newblock Interpretations steered network pruning via amortized inferred
  saliency maps.
\newblock In {\em European Conference on Computer Vision}, pages 278--296.
  Springer, 2022.

\bibitem{gao2022disentangled}
Shangqian Gao, Feihu Huang, Yanfu Zhang, and Heng Huang.
\newblock Disentangled differentiable network pruning.
\newblock In {\em European Conference on Computer Vision}, pages 328--345.
  Springer, 2022.

\bibitem{han2015deep}
Song Han, Huizi Mao, and William~J Dally.
\newblock Deep compression: Compressing deep neural networks with pruning,
  trained quantization and huffman coding.
\newblock {\em arXiv preprint arXiv:1510.00149}, 2015.

\bibitem{han2015learning}
Song Han, Jeff Pool, John Tran, and William Dally.
\newblock Learning both weights and connections for efficient neural network.
\newblock {\em Advances in neural information processing systems}, 28, 2015.

\bibitem{he2016deep}
Kaiming He, Xiangyu Zhang, Shaoqing Ren, and Jian Sun.
\newblock Deep residual learning for image recognition.
\newblock In {\em Proceedings of the IEEE conference on computer vision and
  pattern recognition}, pages 770--778, 2016.

\bibitem{he2022filter}
Zhiqiang He, Yaguan Qian, Yuqi Wang, Bin Wang, Xiaohui Guan, Zhaoquan Gu, Xiang
  Ling, Shaoning Zeng, Haijiang Wang, and Wujie Zhou.
\newblock Filter pruning via feature discrimination in deep neural networks.
\newblock In {\em European Conference on Computer Vision}, pages 245--261.
  Springer, 2022.

\bibitem{hinton2015distilling}
Geoffrey Hinton, Oriol Vinyals, and Jeff Dean.
\newblock Distilling the knowledge in a neural network.
\newblock {\em arXiv preprint arXiv:1503.02531}, 2015.

\bibitem{hoffer2019augment}
Elad Hoffer, Tal Ben-Nun, Itay Hubara, Niv Giladi, Torsten Hoefler, and Daniel
  Soudry.
\newblock Augment your batch: better training with larger batches.
\newblock {\em arXiv preprint arXiv:1901.09335}, 2019.

\bibitem{kang2020operation}
Minsoo Kang and Bohyung Han.
\newblock Operation-aware soft channel pruning using differentiable masks.
\newblock In {\em International Conference on Machine Learning}, pages
  5122--5131. PMLR, 2020.

\bibitem{krizhevsky2009learning}
Alex Krizhevsky, Geoffrey Hinton, et~al.
\newblock Learning multiple layers of features from tiny images.
\newblock 2009.

\bibitem{lecun1989optimal}
Yann LeCun, John Denker, and Sara Solla.
\newblock Optimal brain damage.
\newblock {\em Advances in neural information processing systems}, 2, 1989.

\bibitem{lee2022ensemble}
Seunghyun Lee and Byung~Cheol Song.
\newblock Ensemble knowledge guided sub-network search and fine-tuning for
  filter pruning.
\newblock In {\em European Conference on Computer Vision}, pages 569--585.
  Springer, 2022.

\bibitem{li2016pruning}
Hao Li, Asim Kadav, Igor Durdanovic, Hanan Samet, and Hans~Peter Graf.
\newblock Pruning filters for efficient convnets.
\newblock {\em arXiv preprint arXiv:1608.08710}, 2016.

\bibitem{li2022revisiting}
Yawei Li, Kamil Adamczewski, Wen Li, Shuhang Gu, Radu Timofte, and Luc
  Van~Gool.
\newblock Revisiting random channel pruning for neural network compression.
\newblock In {\em Proceedings of the IEEE/CVF Conference on Computer Vision and
  Pattern Recognition}, pages 191--201, 2022.

\bibitem{li2020group}
Yawei Li, Shuhang Gu, Christoph Mayer, Luc~Van Gool, and Radu Timofte.
\newblock Group sparsity: The hinge between filter pruning and decomposition
  for network compression.
\newblock In {\em Proceedings of the IEEE/CVF conference on computer vision and
  pattern recognition}, pages 8018--8027, 2020.

\bibitem{lin2020hrank}
Mingbao Lin, Rongrong Ji, Yan Wang, Yichen Zhang, Baochang Zhang, Yonghong
  Tian, and Ling Shao.
\newblock Hrank: Filter pruning using high-rank feature map.
\newblock In {\em Proceedings of the IEEE/CVF conference on computer vision and
  pattern recognition}, pages 1529--1538, 2020.

\bibitem{lin2020channel}
Mingbao Lin, Rongrong Ji, Yuxin Zhang, Baochang Zhang, Yongjian Wu, and
  Yonghong Tian.
\newblock Channel pruning via automatic structure search.
\newblock {\em arXiv preprint arXiv:2001.08565}, 2020.

\bibitem{lin2019towards}
Shaohui Lin, Rongrong Ji, Chenqian Yan, Baochang Zhang, Liujuan Cao, Qixiang
  Ye, Feiyue Huang, and David Doermann.
\newblock Towards optimal structured cnn pruning via generative adversarial
  learning.
\newblock In {\em Proceedings of the IEEE/CVF conference on computer vision and
  pattern recognition}, pages 2790--2799, 2019.

\bibitem{liu2017learning}
Zhuang Liu, Jianguo Li, Zhiqiang Shen, Gao Huang, Shoumeng Yan, and Changshui
  Zhang.
\newblock Learning efficient convolutional networks through network slimming.
\newblock In {\em Proceedings of the IEEE international conference on computer
  vision}, pages 2736--2744, 2017.

\bibitem{liu2018rethinking}
Zhuang Liu, Mingjie Sun, Tinghui Zhou, Gao Huang, and Trevor Darrell.
\newblock Rethinking the value of network pruning.
\newblock {\em arXiv preprint arXiv:1810.05270}, 2018.

\bibitem{ma2021non}
Xiaolong Ma, Sheng Lin, Shaokai Ye, Zhezhi He, Linfeng Zhang, Geng Yuan,
  Sia~Huat Tan, Zhengang Li, Deliang Fan, Xuehai Qian, et~al.
\newblock Non-structured dnn weight pruning—is it beneficial in any platform?
\newblock {\em IEEE transactions on neural networks and learning systems},
  33(9):4930--4944, 2021.

\bibitem{molchanov2017variational}
Dmitry Molchanov, Arsenii Ashukha, and Dmitry Vetrov.
\newblock Variational dropout sparsifies deep neural networks.
\newblock In {\em International Conference on Machine Learning}, pages
  2498--2507. PMLR, 2017.

\bibitem{muller2021trivialaugment}
Samuel~G M{\"u}ller and Frank Hutter.
\newblock Trivialaugment: Tuning-free yet state-of-the-art data augmentation.
\newblock In {\em Proceedings of the IEEE/CVF international conference on
  computer vision}, pages 774--782, 2021.

\bibitem{rombach2022high}
Robin Rombach, Andreas Blattmann, Dominik Lorenz, Patrick Esser, and Bj{\"o}rn
  Ommer.
\newblock High-resolution image synthesis with latent diffusion models.
\newblock In {\em Proceedings of the IEEE/CVF conference on computer vision and
  pattern recognition}, pages 10684--10695, 2022.

\bibitem{russakovsky2015imagenet}
Olga Russakovsky, Jia Deng, Hao Su, Jonathan Krause, Sanjeev Satheesh, Sean Ma,
  Zhiheng Huang, Andrej Karpathy, Aditya Khosla, Michael Bernstein, et~al.
\newblock Imagenet large scale visual recognition challenge.
\newblock {\em International journal of computer vision}, 115:211--252, 2015.

\bibitem{shang2022neural}
Haopu Shang, Jia-Liang Wu, Wenjing Hong, and Chao Qian.
\newblock Neural network pruning by cooperative coevolution.
\newblock {\em arXiv preprint arXiv:2204.05639}, 2022.

\bibitem{soulie2016compression}
Guillaume Souli{\'e}, Vincent Gripon, and Ma{\"e}lys Robert.
\newblock Compression of deep neural networks on the fly.
\newblock In {\em Artificial Neural Networks and Machine Learning--ICANN 2016:
  25th International Conference on Artificial Neural Networks, Barcelona,
  Spain, September 6-9, 2016, Proceedings, Part II 25}, pages 153--160.
  Springer, 2016.

\bibitem{szegedy2016rethinking}
Christian Szegedy, Vincent Vanhoucke, Sergey Ioffe, Jon Shlens, and Zbigniew
  Wojna.
\newblock Rethinking the inception architecture for computer vision.
\newblock In {\em Proceedings of the IEEE conference on computer vision and
  pattern recognition}, pages 2818--2826, 2016.

\bibitem{tang2020scop}
Yehui Tang, Yunhe Wang, Yixing Xu, Dacheng Tao, Chunjing Xu, Chao Xu, and Chang
  Xu.
\newblock Scop: Scientific control for reliable neural network pruning.
\newblock {\em Advances in Neural Information Processing Systems},
  33:10936--10947, 2020.

\bibitem{tessier2023convolutional}
Hugo Tessier.
\newblock {\em Convolutional neural networks pruning and its application to
  embedded vision systems}.
\newblock PhD thesis, Ecole nationale sup{\'e}rieure Mines-T{\'e}l{\'e}com
  Atlantique Bretagne Pays de la Loire, 2023.

\bibitem{tessier2022energy}
Hugo Tessier, Vincent Gripon, Mathieu L{\'e}onardon, Matthieu Arzel, David
  Bertrand, and Thomas Hannagan.
\newblock Energy consumption analysis of pruned semantic segmentation networks
  on an embedded gpu.
\newblock In {\em International Conference on System-Integrated Intelligence},
  pages 553--563. Springer, 2022.

\bibitem{tessier2022leveraging}
Hugo Tessier, Vincent Gripon, Mathieu L{\'e}onardon, Matthieu Arzel, David
  Bertrand, and Thomas Hannagan.
\newblock Leveraging structured pruning of convolutional neural networks.
\newblock In {\em 2022 IEEE Workshop on Signal Processing Systems (SiPS)},
  pages 1--6. IEEE, 2022.

\bibitem{tessier2022rethinking}
Hugo Tessier, Vincent Gripon, Mathieu L{\'e}onardon, Matthieu Arzel, Thomas
  Hannagan, and David Bertrand.
\newblock Rethinking weight decay for efficient neural network pruning.
\newblock {\em Journal of Imaging}, 8(3):64, 2022.

\bibitem{touvron2019fixing}
Hugo Touvron, Andrea Vedaldi, Matthijs Douze, and Herv{\'e} J{\'e}gou.
\newblock Fixing the train-test resolution discrepancy.
\newblock {\em Advances in neural information processing systems}, 32, 2019.

\bibitem{yao2019balanced}
Zhuliang Yao, Shijie Cao, Wencong Xiao, Chen Zhang, and Lanshun Nie.
\newblock Balanced sparsity for efficient dnn inference on gpu.
\newblock In {\em Proceedings of the AAAI conference on artificial
  intelligence}, volume~33, pages 5676--5683, 2019.

\bibitem{you2019gate}
Zhonghui You, Kun Yan, Jinmian Ye, Meng Ma, and Ping Wang.
\newblock Gate decorator: Global filter pruning method for accelerating deep
  convolutional neural networks.
\newblock {\em Advances in neural information processing systems}, 32, 2019.

\bibitem{younes2022inter}
Hamoud Younes, Hugo~Le Blevec, Mathieu L{\'e}onardon, and Vincent Gripon.
\newblock Inter-operability of compression techniques for efficient deployment
  of cnns on microcontrollers.
\newblock In {\em International Conference on System-Integrated Intelligence},
  pages 543--552. Springer, 2022.

\bibitem{yu2022topology}
Sixing Yu, Arya Mazaheri, and Ali Jannesari.
\newblock Topology-aware network pruning using multi-stage graph embedding and
  reinforcement learning.
\newblock In {\em International conference on machine learning}, pages
  25656--25667. PMLR, 2022.

\bibitem{yun2019cutmix}
Sangdoo Yun, Dongyoon Han, Seong~Joon Oh, Sanghyuk Chun, Junsuk Choe, and
  Youngjoon Yoo.
\newblock Cutmix: Regularization strategy to train strong classifiers with
  localizable features.
\newblock In {\em Proceedings of the IEEE/CVF international conference on
  computer vision}, pages 6023--6032, 2019.

\bibitem{yvinec2022singe}
Edouard Yvinec, Arnaud Dapogny, Matthieu Cord, and Kevin Bailly.
\newblock Singe: Sparsity via integrated gradients estimation of neuron
  relevance.
\newblock {\em Advances in Neural Information Processing Systems},
  35:35392--35403, 2022.

\bibitem{zhang2017mixup}
Hongyi Zhang, Moustapha Cisse, Yann~N Dauphin, and David Lopez-Paz.
\newblock mixup: Beyond empirical risk minimization.
\newblock {\em arXiv preprint arXiv:1710.09412}, 2017.

\bibitem{zhang2022weighted}
Miao Zhang, Li Wang, David Campos, Wei Huang, Chenjuan Guo, and Bin Yang.
\newblock Weighted mutual learning with diversity-driven model compression.
\newblock {\em Advances in Neural Information Processing Systems},
  35:11520--11533, 2022.

\bibitem{zhong2021revisit}
Shaochen Zhong, Guanqun Zhang, Ningjia Huang, and Shuai Xu.
\newblock Revisit kernel pruning with lottery regulated grouped convolutions.
\newblock In {\em International Conference on Learning Representations}, 2021.

\bibitem{zhong2020random}
Zhun Zhong, Liang Zheng, Guoliang Kang, Shaozi Li, and Yi Yang.
\newblock Random erasing data augmentation.
\newblock In {\em Proceedings of the AAAI conference on artificial
  intelligence}, volume~34, pages 13001--13008, 2020.

\end{thebibliography}
